\documentclass[11pt,a4paper]{article}
\usepackage{authblk}
\usepackage[hyperref]{acl2019}
\usepackage{times}
\usepackage{latexsym}
\usepackage{graphicx}
\usepackage{amsmath}
\usepackage{makecell}
\usepackage{booktabs}
\usepackage{url}
\usepackage{subfigure}
\usepackage{float}
\usepackage{tabularx}

\usepackage{url}

\usepackage{xcolor}
\usepackage{soul}

\aclfinalcopy 

\title{Widening the Representation Bottleneck in Neural Machine Translation with Lexical Shortcuts}

\author[1]{Denis Emelin}
\author[1, 2]{Ivan Titov}
\author[1, 3]{Rico Sennrich}

\affil[1]{University of Edinburgh, Scotland}
\affil[2]{University of Amsterdam, Netherlands}
\affil[3]{University of Zurich, Switzerland}
\affil[ ]{\texttt {D.Emelin@sms.ed.ac.uk}} 
\affil[ ]{\texttt {ititov@inf.ed.ac.uk rico.sennrich@ed.ac.uk}}

\date{}

\begin{document}
\maketitle
\begin{abstract}
  The transformer is a state-of-the-art neural translation model that uses attention to iteratively refine lexical representations with information drawn from the surrounding context. Lexical features are fed into the first layer and propagated through a deep network of hidden layers. We argue that the need to represent and propagate lexical features in each layer limits the model's capacity for learning and representing other information relevant to the task. To alleviate this bottleneck, we introduce gated shortcut connections between the embedding layer and each subsequent layer within the encoder and decoder. This enables the model to access relevant lexical content dynamically, without expending limited resources on storing it within intermediate states. We show that the proposed modification yields consistent improvements over a baseline transformer on standard WMT translation tasks in 5 translation directions (0.9 BLEU on average) and reduces the amount of lexical information passed along the hidden layers. We furthermore evaluate different ways to integrate lexical connections into the transformer architecture and present ablation experiments exploring the effect of proposed shortcuts on model behavior.\footnote{Our code is publicly available to aid the reproduction of the reported results: \url{https://github.com/demelin/transformer\_lexical\_shortcuts}}
\end{abstract}

\section{Introduction}
Since it was first proposed, the transformer model \cite{vaswani2017attention} has quickly established itself as a popular choice for neural machine translation, where it has been found to deliver state-of-the-art results on various translation tasks \cite{bojar-EtAl:2018:WMT1}. Its success can be attributed to the model's high parallelizability allowing for significantly faster training compared to recurrent neural networks \cite{chen2018best}, superior ability to perform lexical disambiguation, and capacity for capturing long-distance dependencies on par with existing alternatives \cite{tang2018self}.

Recently, several studies have investigated the nature of features encoded within individual layers of neural translation models \cite{belinkov2017neural, belinkov2018evaluating}. One central finding reported in this body of work is that, in recurrent architectures, different layers prioritize different information types. As such, lower layers appear to predominantly perform morphological and syntactic processing, whereas semantic features reach their highest concentration towards the top of the layer stack. One necessary consequence of this distributed learning is that different types of information encoded within input representations received by the translation model have to be transported to the layers specialized in exploiting them.

Within the transformer encoder and decoder alike, information exchange proceeds in a strictly sequential manner, whereby each layer attends over the output of the immediately preceding layer, complemented by a shallow residual connection. For input features to be successfully propagated to the uppermost layers, the translation model must therefore store them in its intermediate representations until they can be processed. By retaining lexical content, the model is unable to leverage its full representational capacity for learning new information from other sources, such as the surrounding sentence context. We refer to this limitation as the representation bottleneck.

To alleviate this bottleneck, we propose extending the standard transformer architecture with lexical shortcuts which connect the embedding layer with each subsequent self-attention sub-layer in both encoder and decoder. The shortcuts are defined as gated skip connections, allowing the model to access relevant lexical information at any point, instead of propagating it upwards from the embedding layer along the hidden states. 

We evaluate the resulting model's performance on multiple language pairs and varying corpus sizes, showing a consistent improvement in translation quality over the unmodified transformer baseline. Moreover, we examine the distribution of lexical information across the hidden layers of the transformer model in its standard configuration and with added shortcut connections. The presented experiments provide quantitative evidence for the presence of a representation bottleneck in the standard transformer and its reduction following the integration of lexical shortcuts. 

While our experimental efforts are centered around the transformer, the proposed components are compatible with other multi-layer NMT architectures.

The contributions of our work are as follows:
\begin{enumerate}
\item We propose the use of lexical shortcuts as a simple strategy for alleviating the representation bottleneck in NMT models.
\item We demonstrate significant improvements in translation quality across multiple language pairs as a result of equipping the transformer with lexical shortcut connections.
\item We conduct a series of ablation studies, showing that shortcuts are best applied to the self-attention mechanism in both encoder and decoder.
\item We report a positive impact of our modification on the model's ability to perform word sense disambiguation.
\end{enumerate}

\section{Proposed Method}
\subsection{Background: The transformer}

As defined in \cite{vaswani2017attention}, the transformer is comprised of two sub-networks, the encoder and the decoder. The encoder converts the received source language sentence into a sequence of continuous representations containing translation-relevant features. The decoder, on the other hand, generates the target language sequence, whereby each translation step is conditioned on the encoder's output as well as the translation prefix produced up to that point.

Both encoder and decoder are composed of a series of identical layers. Each encoder layer contains two sub-layers: A self-attention mechanism and a position-wise fully connected feed-forward network. Within the decoder, each layer is extended with a third sub-layer responsible for attending over the encoder's output. In each case, the attention mechanism is implemented as multi-head, scaled dot-product attention, which allows the model to simultaneously consider different context sub-spaces. Additionally, residual connections between layer inputs and outputs are employed to aid with signal propagation.
\begin{figure}[!t]
\centering
\includegraphics[width=7.5cm]{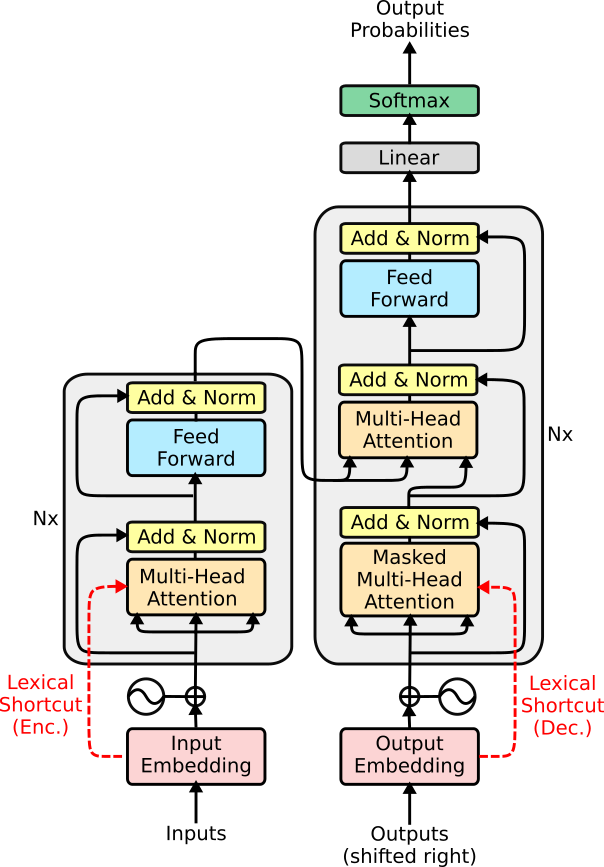}
\caption{Integration of lexical shortcut connections into the overall transformer architecture.}
\label{fig:transformer-shortcut}
\end{figure}

In order for the dot-product attention mechanism to be effective, its inputs first have to be projected into a common representation sub-space. This is accomplished by multiplying the input arrays $H^S$ and $H^T$ by one of the three weight matrices $K$, $V$, and $Q$, as shown in Eqn. \ref{eq:1}-\ref{eq:3}, producing attention keys, values, and queries, respectively. In case of multi-head attention, each head is assigned its own set of keys, values, and queries with the associated learned projection weights.
\begin{align}
Q = W^Q H^S\label{eq:1} \\
K = W^K H^T\label{eq:2} \\
V = W^V H^T\label{eq:3}
\end{align}

In case of encoder-to-decoder attention, $H^T$ corresponds to the final encoder states, whereas $H^S$ is the context vector generated by the preceding self-attention sub-layer. For self-attention, on the other hand, all three operations are given the output of the preceding layer as their input. Eqn. \ref{eq:4} defines attention as a function over the projected representations.  
\begin{equation}\label{eq:4}
Attention(Q, K, V) = \textrm{softmax}(\frac{Q K^T}{\sqrt{d_k}}) V
\end{equation}

To prevent the magnitude of the pre-softmax dot-product from becoming too large, it is divided by the square root of the total key dimensionality $d_k$. Finally, the translated sequence is obtained by feeding the output of the decoder through a softmax activation function and sampling from the produced distribution over target language tokens.

\subsection{Lexical shortcuts}
Given that the attention mechanism represents the primary means of establishing parameterized connections between the different layers within the transformer, it is well suited for the re-introduction of lexical content. We achieve this by adding gated connections between the embedding layer and each subsequent self-attention sub-layer within the encoder and the decoder, as shown in Figure \ref{fig:transformer-shortcut}.

To ensure that lexical features are compatible with the learned hidden representations, the retrieved embeddings are projected into the appropriate latent space, by multiplying them with the layer-specific weight matrices $W^{K^{SC}}_l$ and $W^{V^{SC}}_l$. We account for the potentially variable importance of lexical features by equipping each added connection with a binary gate inspired by the Gated Recurrent Unit \cite{cho2014learning}. Functionally, our lexical shortcuts are equivalent to highway connections of \cite{srivastava2015highway} that span an arbitrary number of intermediate layers.
\begin{align}
K^{SC}_l &= W^{K^{SC}}_l E\label{eq:5} \\
V^{SC}_l &= W^{V^{SC}}_l E\label{eq:6} \\
K_l &= W^K_l H_{l-1}\label{eq:7} \\
V_l &= W^V_l H_{l-1}\label{eq:8}
\end{align}
\begin{align}
r^K_l &= \textrm{sigmoid}(K^{SC}_l + K_l + b^K_l)\label{eq:9} \\
r^V_l &= \textrm{sigmoid}(V^{SC}_l + V_l + b^V_l)\label{eq:10} 
\end{align}
\begin{align}
K'_l &= r^K_l \odot K^{SC}_l + (1 - r^K_l) \odot K_l\label{eq:11} \\
V'_l &= r^V_l \odot V^{SC}_l + (1 - r^V_l) \odot V_l\label{eq:12}
\end{align}

After situating the outputs of the immediately preceding layer $H_{l-1}$ and the embeddings $E$ within a shared representation space (Eqn. \ref{eq:5}-\ref{eq:8}), the relevance of lexical information for the current attention step is estimated by comparing lexical and latent features, followed by the addition of a bias term $b$ (Eqn. \ref{eq:9}-\ref{eq:10}). The respective attention key arrays are denoted as $K^{SC}_l$ and $K_l$, while $V^{SC}_l$ and $V_l$ represent the corresponding value arrays. The result is fed through a sigmoid function to obtain the lexical relevance weight $r$, used to calculate the weighted sum of the two sets of features (Eqn. \ref{eq:11}-\ref{eq:12}), where $\odot$ denotes element-wise multiplication. Next, the key and value arrays $K'_l$ and $V'_l$ are passed to the multi-head attention function as defined in Eqn. \ref{eq:4}, replacing the original $K_l$ and $V_l$. 
\begin{figure}[!t]
\centering
\includegraphics[width=5cm]{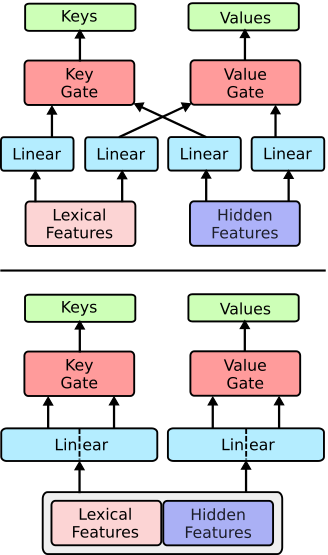}
\caption{Modified attention inputs. Top: lexical shortcuts, bottom: lexical shortcuts + feature-fusion. Dashed lines denote splits along the feature dimension.}
\label{fig:attn-inputs}
\end{figure}

In an alternative formulation of the model, referred to as `feature-fusion' from here on, we concatenate $E$ and $H_{l-1}$ before the initial linear projection, splitting the result in two halves along the feature dimension and leaving the rest of the shortcut definition unchanged\footnote{The feature-fusion mechanism is therefore based on the same principle as the Gated Linear Unit \cite{dauphin2017language}, while utilizing a more expressive gating function.}. This reduces Eqn. \ref{eq:5}-\ref{eq:8} to Eqn. \ref{eq:13}-\ref{eq:14}, and enables the model to select relevant information by directly inter-relating lexical and hidden features. As such, both $K^{SC}_l$ and $K_l$ encode a mixture of embedding and hidden features, as do the corresponding value arrays. While this arguably diminishes the contribution of the gating mechanism towards feature selection, preliminary experiments have shown that replacing gated shortcuts with gate-less residual connections \cite{he2016deep} produces models that fail to converge, characterized by poor training and validation performance. For illustration purposes, figure \ref{fig:attn-inputs} depicts the modified computation path of the lexically-enriched attention key and value vectors.
\begin{align}
K^{SC}_l, K_l  &= W^K_l [E; H_{l-1}]\label{eq:13} \\
V^{SC}_l, V_l  &= W^V_l [E; H_{l-1}]\label{eq:14}
\end{align}

Other than the immediate accessibility of lexical information, one potential benefit afforded by the introduced shortcuts is the improved gradient flow during back-propagation. As noted in \cite{huang2017densely}, the addition of skip connections between individual layers of a deep neural network results in an implicit `deep supervision' effect \cite{lee2015deeply}, which aids the training process. In case of our modified transformer, this corresponds to the embedding layer receiving its learning signal from the model's overall optimization objective as well as from each layer it is connected to, making the model easier to train.

\section{Experiments}
\subsection{Training} \label{results}
To evaluate the efficacy of the proposed approach, we re-implement the transformer model and extend it by applying lexical shortcuts to each self-attention layer in the encoder and decoder. A detailed account of our model configurations, data pre-processing steps, and training setup is given in the appendix (\ref{appendix:training}-\ref{appendix:data}).

\begin{table*}[!h]
\centering
\begin{tabular}{c c cccccc}
    \toprule
& \multicolumn{1}{c}{} & \multicolumn{6}{c}{sacreBLEU} \\
\cmidrule(lr){3-8}
Model & \thead{newstest2014\\(tokenized BLEU)} & \thead{newstest\\2014} & \thead{newstest\\2015} & \thead{newstest\\2016} & \thead{newstest\\2017} & \thead{newstest\\2018} & \thead{test\\mean} \\
    \midrule
transformer-BASE & 27.3 & 25.8 & 28.5 & 33.2 & 27.3 & 40.4 & 31.0 \\
    \midrule
+ lexical shortcuts & 27.6 & 26.1 & 29.5 & 33.3 & 27.5 & 41.1 & 31.5  \\
    \midrule
+ feature-fusion & \textbf{28.3} & \textbf{26.8} & \textbf{29.9} & \textbf{34.0} & \textbf{27.7} & \textbf{41.6} & \textbf{32.0} \\
    \midrule
    \midrule
transformer-BIG & 28.7 & 27.2 & 30.1 & \textbf{34.0} & 28.1 & 41.3 & 32.1 \\ 
    \midrule
\thead{+ lexical shortcuts\\+ feature-fusion} & \textbf{29.4} & \textbf{27.8} & \textbf{30.3} & 33.2 & \textbf{28.4} & 41.3 & \textbf{32.2} \\
\bottomrule
\end{tabular}
\caption{BLEU scores for the EN$\rightarrow$DE news translation task.}
\label{tab:en->de}
\end{table*}

\begin{table*}[!t]
\centering
\begin{tabular}{c cc cc cc cc}
\toprule
& \multicolumn{2}{c}{DE$\rightarrow$EN} & \multicolumn{2}{c}{EN$\rightarrow$RU} & \multicolumn{2}{c}{EN$\rightarrow$CS} & \multicolumn{2}{c}{EN$\rightarrow$FI} \\
\cmidrule(lr){2-3}
\cmidrule(lr){4-5}
\cmidrule(lr){6-7}
\cmidrule(lr){8-9}
Model & \thead{newstest\\2014} & \thead{newstest\\2017} & \thead{newstest\\2017} & \thead{newstest\\2018} & \thead{newstest\\2015} & \thead{newstest\\2018} & \thead{newstest\\2015} & \thead{newstest\\2018} \\
    \midrule
transformer-BASE & 31.1 & 32.3 & 27.9 & 24.2 & 23.4 & 21.1 & 18.7 & 14.0\\ 
    \midrule
+ lexical shortcuts & 31.3 & 32.3 & 28.4 & 24.9 & 24.1 & 21.4 & 19.5 & 14.5 \\
    \midrule
+ feature-fusion & \textbf{31.7} & \textbf{32.9} & \textbf{28.9} & \textbf{25.3} & \textbf{24.3} & \textbf{21.6} & \textbf{19.8} & \textbf{14.8} \\
\bottomrule
\end{tabular}
\caption{Effect of lexical shortcuts on translation quality for different language pairs, as measured by sacreBLEU.}
\label{tab:multi-lang-bleu}
\end{table*}

\subsection{Data}
We investigate the potential benefits of lexical shortcuts on 5 WMT translation tasks: German $\rightarrow$ English (DE$\rightarrow$EN), English $\rightarrow$ German (EN$\rightarrow$DE), English $\rightarrow$ Russian (EN$\rightarrow$RU), English $\rightarrow$ Czech (EN$\rightarrow$CS), and English $\rightarrow$ Finnish (EN$\rightarrow$FI). Our choice is motivated by the differences in size of the training corpora as well as by the typological diversity of the target languages. 

To make our findings comparable to related work, we train EN$\leftrightarrow$DE models on the WMT14 news translation data which encompasses $\sim$4.5M sentence pairs. EN$\rightarrow$RU models are trained on the WMT17 version of the news translation task, consisting of $\sim$24.8M sentence pairs. For EN$\rightarrow$CS and EN$\rightarrow$FI, we use the respective WMT18 parallel training corpora, with the former containing $\sim$50.4M and the latter $\sim$3.2M sentence pairs. We do not employ backtranslated data in any of our experiments to further facilitate comparisons to existing work.

Throughout training, model performance is validated on newstest2013 for EN$\leftrightarrow$DE, newstest2016 for EN$\rightarrow$RU, and on newstest2017 for EN$\rightarrow$CS and EN$\rightarrow$FI. Final model performance is reported on multiple tests sets from the news domain for each direction.

\subsection{Translation performance}
The results of our translation experiments are summarized in Tables \ref{tab:en->de}-\ref{tab:multi-lang-bleu}. To ensure their comparability, we evaluate translation quality using sacreBLEU \cite{post2018call}. As such, our baseline performance diverges from that reported in \cite{vaswani2017attention}. We address this by evaluating our EN$\rightarrow$DE models using the scoring script from the tensor2tensor toolkit\footnote{\url{https://github.com/tensorflow/tensor2tensor/blob/master/\\tensor2tensor/utils/get\_ende\_bleu.sh}} \cite{vaswani2018tensor2tensor} on the tokenized model output, and list the corresponding BLEU scores in the first column of Table \ref{tab:en->de}. 

Our evaluation shows that the introduction of lexical shortcuts consistently improves translation quality of the transformer model across different test-sets and language pairs, outperforming transformer-BASE by 0.5 BLEU on average. With feature-fusion, we see even stronger improvements, yielding total performance gains over transformer-BASE of up to 1.4 BLEU for EN$\rightarrow$DE (averaging to 1.0), and 0.8 BLEU on average for the other 4 translation directions. We furthermore observe that the relative improvements from the addition of lexical shortcuts are substantially smaller for transformer-BIG compared to transformer-BASE. One potential explanation for this drop in efficacy is the increased size of latent representations the wider model is able to learn, which we discuss in section \ref{bottleneck}.

Furthermore, it is worth noting that transformer-BASE equipped with lexical connections performs comparably to the standard transformer-BIG, despite containing fewer than half of its parameters and being only marginally slower to train than our unmodified transformer-BASE implementation. A detailed overview of model sizes and training speed is provided in the supplementary material (\ref{appendix:training}).

Concerning the examined language pairs, the average increase in BLEU is highest for EN$\rightarrow$RU (1.1 BLEU) and lowest for DE$\rightarrow$EN (0.6 BLEU). A potential explanation for why this is the case could be the difference in language topology. Of all target languages we consider, English has the least complex morphological system where individual words carry little inflectional information, which stands in stark contrast to a highly inflectional language with a flexible word order such as Russian. It is plausible that lexical shortcuts are especially important for translation directions where the target language is morphologically rich and where the surrounding context is essential to accurately predicting a word's case and agreement. With the proposed shortcuts in place, the transformer has more capacity for modeling such context information.

To investigate the role of lexical connections within the transformer, we conduct a thorough examination of our models' internal representations and learning behaviour. The following analysis is based on models utilizing lexical shortcuts with feature-fusion, due to its superior performance.

\section{Analysis}
\subsection{Representation bottleneck} \label{bottleneck}
The proposed approach is motivated by the hypothesis that the transformer retains lexical features within its individual layers, which limits its capacity for learning and representing other types of relevant information. Direct connections to the embedding layer alleviate this by providing the model with access to lexical features at each processing step, reducing the need for propagating them along hidden states. To investigate whether this is indeed the case, we perform a probing study, where we estimate the amount of lexical content present within each encoder and decoder state.

We examine the internal representations learned by our models by modifying the probing technique introduced in \cite{belinkov2017neural}. Specifically, we train a separate lexical classifier for each layer of a frozen translation model. Each classifier receives hidden states extracted from the respective transformer layer\footnote{We treat the output of the feed-forward sub-layer as that layer's hidden state.} and is tasked with reconstructing the sub-word corresponding to the position of each hidden state. Encoder-specific classifiers learn to reconstruct sub-words in the source sentence, whereas classifiers trained on decoder states are trained to reconstruct target sub-words.

\begin{figure}[!t]
\centering
\includegraphics[width=5.5cm]{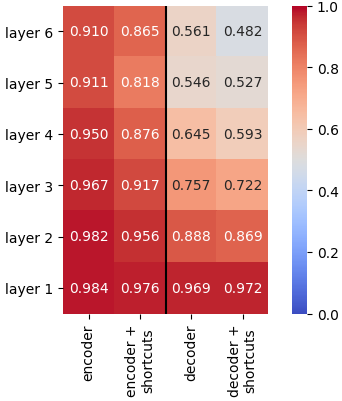}
\caption{Layer-wise lexical probe accuracy measured on transformer-BASE for EN$\rightarrow$DE (newstest2014).}
\label{fig:probe-ende-same}
\end{figure}

\begin{figure}[!t]
\centering
\includegraphics[width=5.5cm]{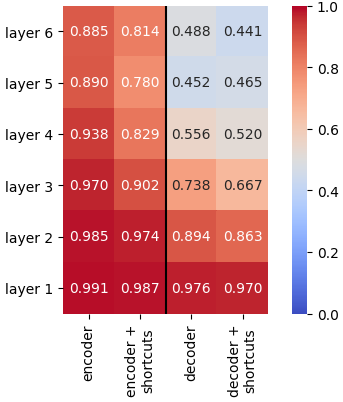}
\caption{Layer-wise lexical probe accuracy measured on transformer-BASE for EN$\rightarrow$RU (newstest2017).}
\label{fig:probe-enru-same}
\end{figure}

The accuracy of each layer-specific classifier on a withheld test set is assumed to be indicative of the lexical content encoded by the corresponding transformer layer. We expect classification accuracy to be high if the evaluated representations predominantly store information propagated upwards from the embeddings at the same position and to decrease proportionally to the amount of information drawn from the surrounding sentence context. Figures \ref{fig:probe-ende-same} and \ref{fig:probe-enru-same} offer a side-by-side comparison of the accuracy scores obtained for each layer of the base transformer and its variant equipped with lexical shortcut connections.

Based on the observed classification results, it appears that immediate access to lexical information does indeed alleviate the representation bottleneck by reducing the extent to which (sub-)word-level content is retained across encoder and decoder layers. By introducing shortcut connections, we effectively reduce the amount of lexical information the model retains within its intermediate states, thereby increasing its capacity for exploiting sentence context. The effect is consistent across multiple language pairs, supporting its generality. Additionally, to examine whether lexical retention depends on the specific properties of the input tokens, we track classification accuracy conditioned on part-of-speech tags and sub-word frequencies. While we do not discover a pronounced effect of either category on classification accuracy, we present a summary of our findings as part of the supplementary material for future reference (\ref{appendix:probes}).

Another observation arising from the probing analysis is that the decoder retains fewer lexical features beyond its initial layers than the encoder. This may be due to the decoder having to represent information it receives from the encoder in addition to target-side content, necessitating a lower rate of lexical feature retention. Even so, by adding shortcut connections we can increase the dissimilarity between the embedding layer and the subsequent layers of the decoder, indicating a noticeable reduction in the retention and propagation of lexical features along the decoder's hidden states.

A similar trend can be observed when evaluating layer similarity directly, which we accomplish by calculating the cosine similarity between the embeddings and the hidden states of each transformer layer. Echoing our findings so far, the addition of lexical shortcuts reduces layer similarity relative to the baseline transformer for both encoder and decoder. The corresponding visualizations are also provided in the appendix (\ref{appendix:probes}).

Overall, the presented analysis supports the existence of a representation bottleneck in NMT models as one potential explanation for the efficacy of the proposed lexical shortcut connections. 

\begin{table}[!h]
\centering
\resizebox{\linewidth}{!}{%
\begin{tabular}{c c c c}
    \toprule
Model & \thead{newstest\\2017} & \thead{newstest\\2018} & \thead{test\\mean} \\
    \midrule
transformer-SMALL & 25.2 & 37.0 & 28.6 \\
    \midrule
+ lexical shortcuts & 25.7 & 38.0 & 29.3 \\
    \midrule
+ feature-fusion & \textbf{25.7} & \textbf{38.5} & \textbf{29.6} \\
\bottomrule
\end{tabular}}
\caption{sacreBLEU scores for small EN$\rightarrow$DE models; `test mean' denotes the average of test-sets in table (1).}
\label{tab:small-en->de}
\end{table}

\subsection{Model size}
Next, we investigate the interaction between the number of model parameters and improvements in translation quality afforded by the proposed lexical connections. Following up on findings presented in section \ref{results}, we hypothesize that the benefit of lexical shortcuts diminishes once the model's capacity is sufficiently large. To establish whether this decline in effectiveness is gradual, we scale down the standard transformer, halving the size of its embeddings, hidden states, and feed-forward sub-layers. Table \ref{tab:small-en->de} shows that, on average, quality improvements are comparable for the small and standard transformer (1.0 BLEU for both), which is in contrast to our observations for transformer-BIG. One explanation is that given sufficient capacity, the model is capable of accommodating the upward propagation of lexical features without having to neglect other sources of information. However, as long as the model's representational capacity is within certain limits, the effect of lexical shortcuts remains comparable across a range of model sizes. With this in mind, the exact interaction between model scale and the types of information encoded in its hidden states remains to be fully explored. We leave a more fine-grained examination of this relationship to future research. 

\subsection{Shortcut variants} \label{variants}
Until now, we focused on applying shortcuts to self-attention as a natural re-entry point for lexical content. However, previous studies suggest that providing the decoder with direct access to source sentences can improve translation adequacy, by conditioning decoding on relevant source tokens \cite{kuang2017attention, nguyen2017improving}.

To investigate whether the proposed method can confer a similar benefit to the transformer, we apply lexical shortcuts to decoder-to-encoder attention, replacing or adding to shortcuts feeding into self-attention. Formally, this equates to fixing $E$ to $E^{enc}$ in Eqn. \ref{eq:5}-\ref{eq:6} and can be regarded as a variant of source-side bridging proposed by \cite{kuang2017attention}. As Table \ref{tab:shortcut-variants} shows, while integrating shortcut connections into the decoder-to-encoder attention improves upon the base transformer, the improvement is smaller than when we modify self-attention.
Interestingly, combining both methods yields worse translation quality than either one does in isolation, indicating that the decoder is unable to effectively consolidate information from both source and target embeddings, which negatively impacts its learned latent representations. We therefore conclude that lexical shortcuts are most beneficial to self-attention.

\begin{table}[!t]
\centering
\resizebox{\linewidth}{!}{%
\begin{tabular}{c c c c}
    \toprule
Model & \thead{newstest\\2017} & \thead{newstest\\2018} & \thead{test\\mean} \\
    \midrule
transformer-BASE & 27.3 & 40.4 & 31.0 \\
    \midrule
+ self-attn. shortcuts & \textbf{27.7} & \textbf{41.6} & \textbf{32.0} \\
    \midrule
    \midrule
dec-to-enc shortcuts & 27.6 & 40.7 & 31.5 \\
    \midrule
+ self-attn. shortcuts & 27.7 & 40.5 & 31.4 \\ 
    \midrule
    \midrule
non-lexical shortcuts & 27.1 & 40.6 & 31.3 \\
\bottomrule
\end{tabular}}
\caption{sacreBLEU for shortcut variants of EN$\rightarrow$DE models; `test mean' averages over test-sets in table (1).}
\label{tab:shortcut-variants}
\end{table}

\begin{figure}[!t]
\centering
\includegraphics[width=7.5cm]{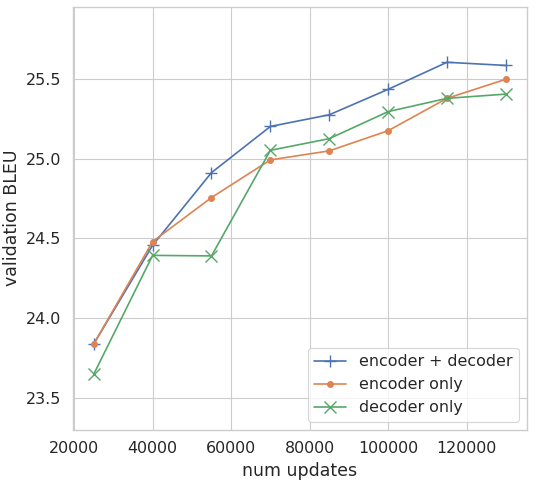}
\caption{Effect of disabling shortcuts in either sub-network on validation BLEU.}
\label{fig:ablations-bleu}
\end{figure}

A related question is whether the encoder and decoder benefit from the addition of lexical shortcuts to self-attention equally. We explore this by disabling shortcuts in either sub-network and comparing the so obtained translation models to one with intact connections. Figure \ref{fig:ablations-bleu} illustrates that best translation performance is obtained by enabling shortcuts in both encoder and decoder. This also improves training stability, as compared to the decoder-only ablated model. The latter may be explained by our use of tied embeddings which receive a stronger training signal from shortcut connections due to `deep supervision', as this may bias learned embeddings against the sub-network lacking improved lexical connectivity.

While adding shortcuts improves translation quality, it is not obvious whether this is predominantly due to improved accessibility of lexical content, rather than increased connectivity between network layers, as suggested in \cite{dou2018exploiting}. To isolate the importance of lexical information, we equip the transformer with non-lexical shortcuts connecting each layer $n$ to layer $n-2$, e.g. layer 6 to layer 4.\footnote{The first layer is connected to the embedding layer, as there is no further antecedent.} As a result, the number of added connections and parameters is kept identical to lexical shortcuts, whereas lexical accessibility is disabled, allowing for minimal comparison between the two configurations. Test-BLEU reported in Table \ref{tab:shortcut-variants} suggests that while non-lexical shortcuts improve over the baseline model, they perform noticeably worse than lexical connections. Therefore, the increase in translation quality associated with lexical shortcuts is not solely attributable to a better signal flow or the increased number of trainable parameters.

\subsection{Word-sense disambiguation}
Beyond the effects of lexical shortcuts on the transformer's learning dynamics, we are interested in how widening the representation bottleneck affects the properties of the produced translations. One challenging problem in translation which intuitively should benefit from the model's increased capacity for learning information drawn from sentence context is word-sense disambiguation. 

We examine whether the addition of lexical shortcuts aids disambiguation by evaluating our trained DE$\rightarrow$EN models on the \textit{ContraWSD} corpus \cite{rios2017improving}. The contrastive dataset is constructed by paring source sentences with multiple translations, varying the translated sense of selected source nouns between translation candidates. A competent model is expected to assign a higher probability to the translation hypothesis containing the appropriate word-sense.

While the standard transformer offers a strong baseline for the disambiguation task, we nonetheless observe improvements after adding direct connections to the embedding layers. Specifically, our baseline model reaches an accuracy of 88.8\%, which improves to 89.5\% with lexical shortcuts.

\section{Related Work}
Within recent literature, several strategies for altering the flow of information within the transformer have been proposed, including adaptive model depth \cite{dehghani2018universal}, layer-wise transparent attention \cite{bapna2018training}, and dense inter-layer connections \cite{dou2018exploiting}. Our investigation bears strongest resemblance to the latter work, by introducing additional connectivity to the model. However, rather than establishing new connections between layers indiscriminately, we explicitly seek to facilitate the accessibility of lexical features across network layers. As a result, our proposed shortcuts remain sparse, while performing comparably to their best, more elaborate strategies that rely on multi-layer attention and hierarchical state aggregation.

Likewise, studies investigating the role of lexical features in NMT are highly relevant to our work. Among them, \cite{nguyen2017improving} note that improving accessibility of source words in the decoder benefits translation quality for low-resource settings. In a similar vein, \cite{wu2018word} attend both encoder hidden states and source embeddings as part of decoder-to-encoder attention, while \cite{kuang2017attention} provide the decoder-to-encoder attention mechanism with improved access to source word representations. We have found a variant of the latter method, which we adapted to the Transformer architecture, to be less effective than applying lexical shortcuts to self-attention, as discussed in section \ref{variants}.

Another line of research from which we draw inspiration concerns itself with the analysis of the internal dynamics and learned representations within deep neural networks \cite{karpathy2015visualizing, shi2016does, qian2016analyzing}. Here, \cite{belinkov2017neural} and \cite{belinkov2018evaluating} serve as our primary points of reference by offering a thorough and principled investigation of the extent to which neural translation models are capable of learning linguistic properties from raw text. 

Our view of the transformer as a model learning to refine input representations through the repeated application of attention is consistent with the iterative estimation paradigm introduced in \cite{greff2016highway}. According to this interpretation, given a stack of connected layers sharing the same dimensionality and interlinked through highway or residual connections, the initial layer generates a rough version of the stack's final output, which is iteratively refined by successive layers, e.g. by enriching localized features with information drawn from the surrounding context. The results of our probing studies support this analysis, further suggesting that different layers not only refine input features but also learn entirely new information given sufficient capacity, as evidenced by the decrease in similarity between embeddings and hidden states with increasing model depth.

\section{Conclusion}
In this paper, we have proposed a simple yet effective method for widening the representation bottleneck in the transformer by introducing lexical shortcuts. Our modified models achieve up to 1.4 BLEU (0.9 BLEU on average) improvement on 5 standard WMT datasets, at a small cost in computing time and model size. Our analysis suggests that lexical connections are useful to both encoder and decoder, and remain effective when included in smaller models. Moreover, the addition of shortcuts noticeably reduces the similarity of hidden states to the initial embeddings, indicating that dynamic lexical access aids the network in learning novel, diverse information. We also performed ablation studies comparing different shortcut variants and demonstrated that one effect of lexical shortcuts is an improved WSD capability.

The presented findings offer new insights into the nature of information encoded by the transformer layers, supporting the iterative refinement view of feature learning. In future work, we intend to explore other ways to better our understanding of the refinement process and to help translation models learn more diverse and meaningful internal representations.

\section{Acknowledgments}
Ivan Titov is supported by the European Union's Horizon 2020 research and innovation programme under grant agreement No 825299 (GoURMET). Computing resources were provided by the Alan Turing Institute under the EPSRCgrant EP/N510129/1.

\bibliography{acl2019}
\bibliographystyle{acl_natbib}

\clearpage
\appendix
\section{Supplementary Material}
\subsection{Training details}
\label{appendix:training}

The majority of our experiments is conducted using the transformer-BASE configuration, with the number of encoder and decoder layers set to 6 each, embedding and attention dimensionality to 512, number of attention heads to 8, and feed-forward sub-layer dimensionality to 2048. We tie the encoder embedding table with the decoder embedding table and the pre-softmax projection matrix to speed up training, following \cite{press2016using}. All trained models are optimized using Adam \cite{kingma2014adam} adhering to the learning rate schedule described in \cite{vaswani2017attention}. We set the number of warm-up steps to 4000 for the baseline model, increasing it to 6000 and 8000 when adding lexical shortcuts and feature-fusion, respectively, so as to accommodate the increase in parameter size. 

We also evaluate the effect of lexical shortcuts on the larger transformer-BIG model, limiting this set of experiments to EN$\rightarrow$DE due to computational constraints. Here, the baseline model employs 16 attention heads, with attention, embedding, and feed-forward dimensions doubled to 1024, 1024, and 4096. Warm-up period for all big models is 16,000 steps. For our probing experiments, the classifiers used are simple feed-forward networks with a single hidden layer consisting of 512 units, dropout \cite{srivastava2014dropout} with $p$ = 0.5, and a ReLU non-linearity. In all presented experiments, we employ beam search during decoding, with beam size set to 16.

\begin{table}[!h]
\centering
\resizebox{\linewidth}{!}{%
\begin{tabular}{c c c}
    \toprule
Model & \# Parameters & Words / sec.  \\
    \midrule
transformer-BASE & 65,166K & 29,698 \\
    \midrule
+ lexical shortcuts & 71,470K & 26,423 \\
    \midrule
+ feature-fusion & 84,053K & 23,601 \\
    \midrule
    \midrule
transformer-BIG & 218,413K & 10,215 \\ 
    \midrule
+ feature-fusion & 293,935K & 6,769 \\
\bottomrule
\end{tabular}}
\caption{Model size and training speed of the compared transformer variants.}
\label{tab:size-and-speed}
\end{table}

All models are trained concurrently on four Nvidia P100 Tesla GPUs using synchronous data parallelization. Delayed optimization \cite{saunders2018multi} is employed to simulate batch sizes of 25,000 tokens, to be consistent with \cite{vaswani2017attention}. Each transformer-BASE model is trained for a total of 150,000 updates, while our transformer-BIG experiments are stopped after 300,000 updates. Validation is performed every 4000 steps, as is check-pointing. Training base models takes $\sim$43 hours, while the addition of shortcut connections increases training time up to $\sim$46 hours ($\sim$50 hours with feature-fusion). Table \ref{tab:size-and-speed} details the differences in parameter size and training speed for the different transformer configurations. Parameters are given in thousands, while speed is averaged over the entire training duration.

Validation-BLEU is calculated using multi-bleu-detok.pl\footnote{\url{https://github.com/moses-smt/mosesdecoder/blob/master/scripts/generic/multi-bleu-detok.perl}} on a reference which we pre- and post-process following the same steps as for the models' inputs and outputs. All reported test-BLEU scores were obtained by averaging the final 5 checkpoints for transformer-BASE and final 16 for transformer-BIG.

\subsection{Data pre-processing}
\label{appendix:data}
We tokenize, clean, and truecase each training corpus using scripts from the Moses toolkit\footnote{\url{https://github.com/moses-smt/mosesdecoder}}, and apply byte-pair encoding \cite{sennrich2015neural} to counteract the open vocabulary issue. Cleaning is skipped for validation and test sets. For EN$\leftrightarrow$DE and EN$\rightarrow$RU we limit the number of BPE merge operations to 32,000 and set the vocabulary threshold to 50. For EN$\rightarrow$CS and EN$\rightarrow$FI, the number of merge operations is set to 89,500 with a vocabulary threshold of 50, following \cite{haddow-EtAl:2018:WMT}\footnote{We do not use synthetic data, which makes our results not directly comparable to theirs.}. In each case, the BPE vocabulary is learned jointly over the source and target language, which necessitated an additional transliteration step for the pre-processing of Russian data\footnote{We used `Lingua Translit' for this purpose: \url{https://metacpan.org/release/Lingua-Translit}}. 

\subsection{Probing studies}
\label{appendix:probes}
\begin{figure}[!h]
\centering
\includegraphics[width=4cm]{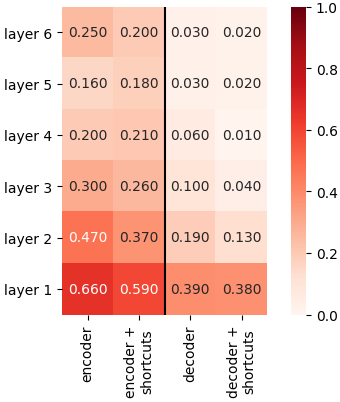}
\caption{Cosine similarity measured on transformer-BASE for EN$\rightarrow$DE (newstest2014).}
\label{fig:probe-ende-sim}
\end{figure}

\begin{figure}[!h]
\centering
\includegraphics[width=4cm]{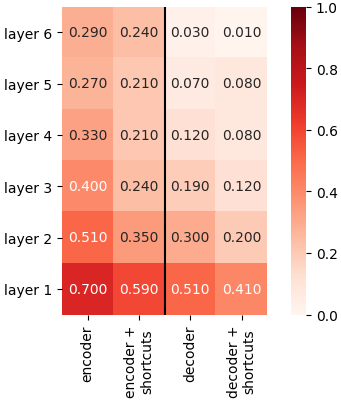}
\caption{Cosine similarity measured on transformer-BASE for EN$\rightarrow$RU (newstest2017).}
\label{fig:probe-enru-sim}
\end{figure}

Cosine similarity scores between the embedding layer and each successive layer in transformer-BASE and its variant equipped with lexical shortcuts are summarized in Figures \ref{fig:probe-ende-sim}-\ref{fig:probe-enru-sim}. 

\begin{figure}[!h]
\centering
\includegraphics[width=8cm]{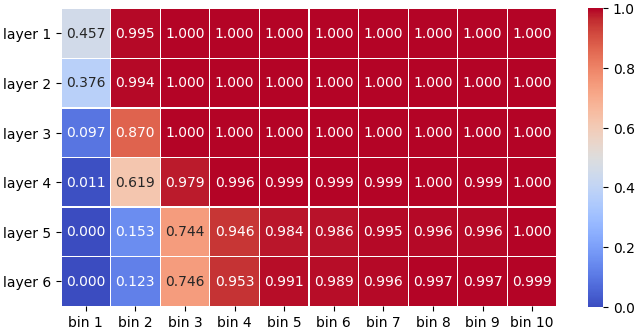}
\caption{Frequency-based classification accuracy on states from the EN$\rightarrow$DE encoder.}
\label{fig:ende-enc-freq}
\end{figure}

\begin{figure}[!h]
\centering
\includegraphics[width=8cm]{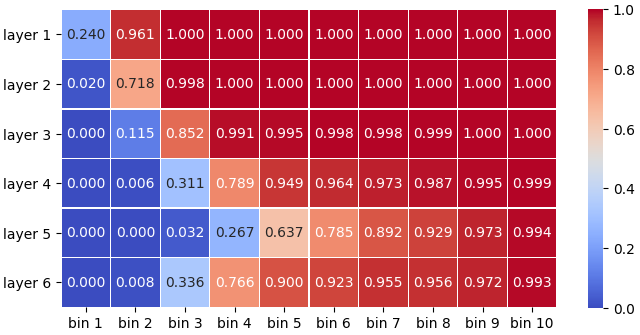}
\caption{Frequency-based classification accuracy on states from the EN$\rightarrow$DE encoder + lexical shortcuts.}
\label{fig:ende-enc-freq-sc}
\end{figure}

\begin{figure}[!h]
\centering
\includegraphics[width=8cm]{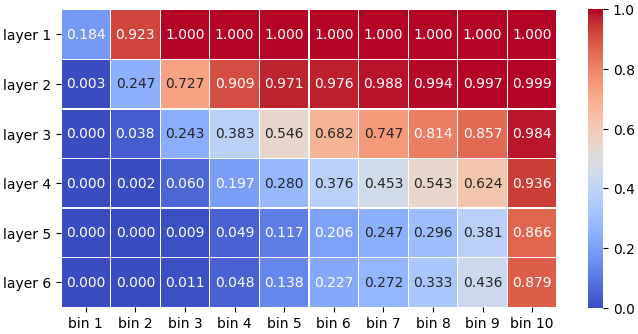}
\caption{Frequency-based classification accuracy on states from the EN$\rightarrow$DE decoder.}
\label{fig:ende-dec-freq}
\end{figure}

\begin{figure}[!h]
\centering
\includegraphics[width=8cm]{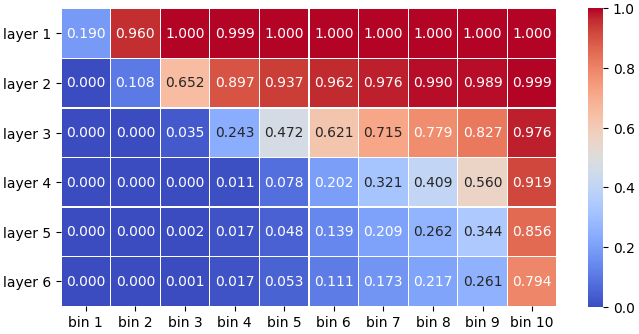}
\caption{Frequency-based classification accuracy on states from the EN$\rightarrow$DE decoder + lexical shortcuts.}
\label{fig:ende-dec-freq-sc}
\end{figure}

\begin{figure}[!h]
\centering
\includegraphics[width=8cm]{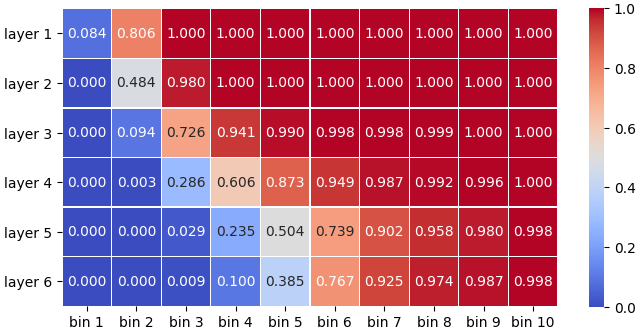}
\caption{Frequency-based classification accuracy on states from the EN$\rightarrow$RU encoder.}
\label{fig:enru-enc-freq}
\end{figure}

\begin{figure}[!h]
\centering
\includegraphics[width=8cm]{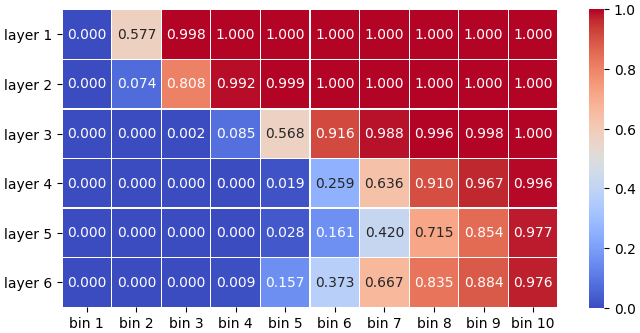}
\caption{Frequency-based classification accuracy on states from the EN$\rightarrow$RU encoder + lexical shortcuts.}
\label{fig:enru-enc-freq-sc}
\end{figure}

\begin{figure}[!h]
\centering
\includegraphics[width=8cm]{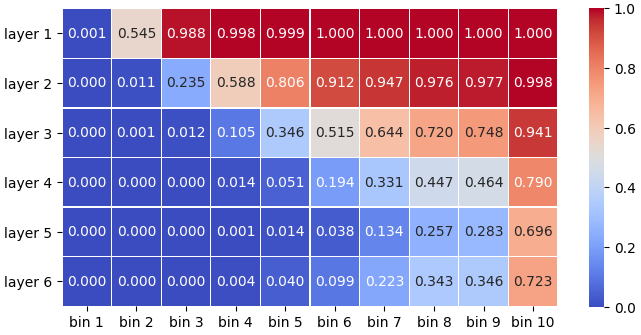}
\caption{Frequency-based classification accuracy on states from the EN$\rightarrow$RU decoder.}
\label{fig:enru-dec-freq}
\end{figure}

\begin{figure}[!h]
\centering
\includegraphics[width=8cm]{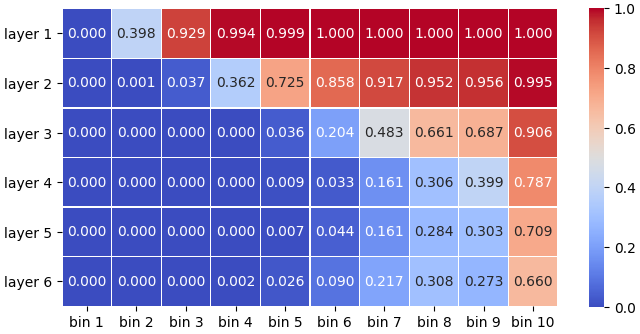}
\caption{Frequency-based classification accuracy on states from the EN$\rightarrow$RU decoder + lexical shortcuts.}
\label{fig:enru-dec-freq-sc}
\end{figure}

\begin{figure}[!h]
\centering
\includegraphics[width=4cm]{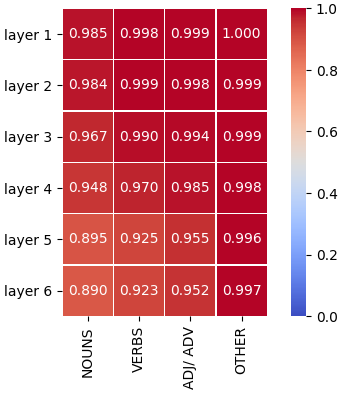}
\caption{POS-based classification accuracy on states from the EN$\rightarrow$DE encoder.}
\label{fig:ende-enc-pos}
\end{figure}

\begin{figure}[!h]
\centering
\includegraphics[width=4cm]{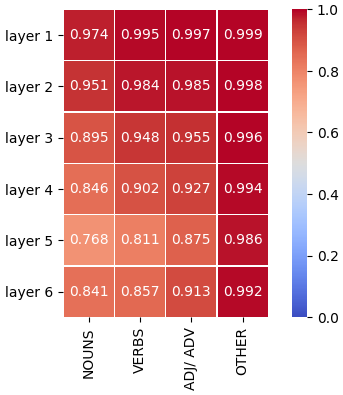}
\caption{POS-based classification accuracy on states from the EN$\rightarrow$DE encoder + lexical shortcuts.}
\label{fig:ende-enc-pos-sc}
\end{figure}

\begin{figure}[!h]
\centering
\includegraphics[width=4cm]{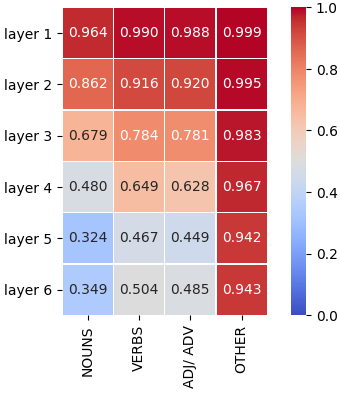}
\caption{POS-based classification accuracy on states from the EN$\rightarrow$DE decoder.}
\label{fig:ende-dec-pos}
\end{figure}

\begin{figure}[!h]
\centering
\includegraphics[width=4cm]{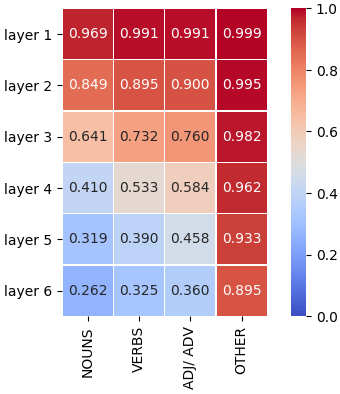}
\caption{POS-based classification accuracy on states from the EN$\rightarrow$DE decoder + lexical shortcuts.}
\label{fig:ende-dec-pos-sc}
\end{figure}

\begin{figure}[!h]
\centering
\includegraphics[width=4cm]{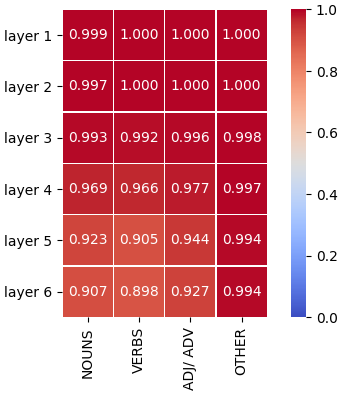}
\caption{POS-based classification accuracy on states from the EN$\rightarrow$RU encoder.}
\label{fig:enru-enc-pos}
\end{figure}

\begin{figure}[!h]
\centering
\includegraphics[width=4cm]{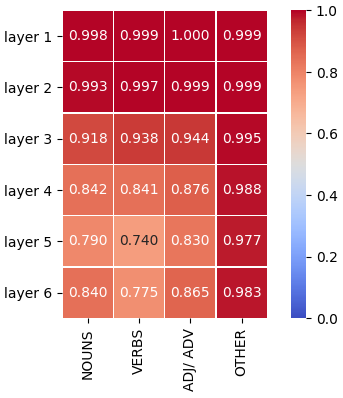}
\caption{POS-based classification accuracy on states from the EN$\rightarrow$RU encoder + lexical shortcuts.}
\label{fig:enru-enc-pos-sc}
\end{figure}

\begin{figure}[!h]
\centering
\includegraphics[width=4cm]{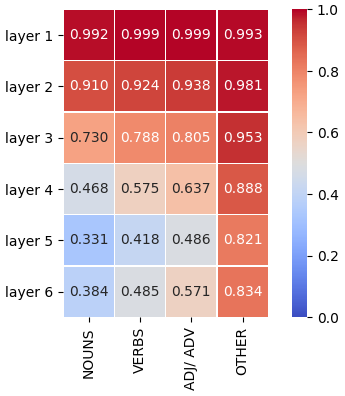}
\caption{POS-based classification accuracy on states from the EN$\rightarrow$RU decoder.}
\label{fig:enru-dec-pos}
\end{figure}

\begin{figure}[!h]
\centering
\includegraphics[width=4cm]{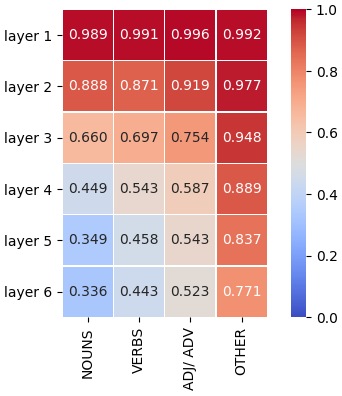}
\caption{POS-based classification accuracy on states from the EN$\rightarrow$RU decoder + lexical shortcuts.}
\label{fig:enru-dec-pos-sc}
\end{figure}

For our fine-grained probing studies, we evaluated classification accuracy conditioned of part-of-speech tags and sub-word frequencies. For the former, we first parse our test-sets with TreeTagger \cite{schmid1999improvements}, projecting tags onto the constituent sub-words of each annotated word. For frequency-based evaluation, we divide sub-words into ten equally-sized frequency bins, with bin 1 containing the least frequent sub-words and bin 10 containing the most frequent ones. We do not observe any immediately obvious, significant effects of either POS or frequency on the retention of lexical features. While classification accuracy is notably low for infrequent sub-words, this can be attributed to the limited occurrence of the corresponding transformer states in the classifier's training data. Evaluation for EN$\rightarrow$DE models is done on newstest2014, while newstest2017 is used for EN$\rightarrow$RU models. Figures \ref{fig:ende-enc-freq}-\ref{fig:enru-dec-freq-sc} present results for the frequency-based classification. Accuracy scores conditioned on POS tags are visualized in Figures \ref{fig:ende-enc-pos}-\ref{fig:enru-dec-pos-sc}.

We also investigated the activation patterns of the lexical shortcut gates. However, despite their essential status for the successful training of transformer variants equipped with lexical connections, we were unable to discern any distinct patterns in the activations of the individual gates, which tend to prioritize lexical and hidden features to an equal degree regardless of training progress or (sub-)word characteristics.

\newpage

\end{document}